\documentclass[10pt,twocolumn,letterpaper]{article}

\usepackage{cvpr}
\usepackage{times}
\usepackage{epsfig}
\usepackage{graphicx}
\usepackage{amsmath}
\usepackage{amssymb}
\usepackage{subfigure}

\usepackage[pagebackref=true,breaklinks=true,letterpaper=true,colorlinks,bookmarks=false]{hyperref}

\cvprfinalcopy 

\def\cvprPaperID{1114} 

\ifcvprfinal\fi
\begin{document}

\title{Do More Drops in Pool$_5$ Feature Maps for Better Object Detection}

\author{Zhiqiang Shen\\
Fudan University\\
{\tt\small zhiqiangshen13@fudan.edu.cn}
\and
Xiangyang Xue\\
Fudan University\\
{\tt\small xyxue@fudan.edu.cn}
}

\maketitle



 
\begin{abstract}
Deep Convolutional Neural Networks (CNNs) have gained great success in image classification and object detection. In these fields, the outputs of all layers of CNNs are usually considered as a high dimensional feature vector extracted from an input image and the correspondence between finer level feature vectors and  concepts that the input image contains is all-important. However, fewer studies focus on this deserving issue. On considering the correspondence, we propose a novel approach which generates an edited version for each original CNN feature vector by applying the maximum entropy principle to abandon particular vectors. These selected vectors correspond to the unfriendly concepts in each image category. The classifier trained from merged feature sets can significantly improve model generalization of individual categories when training data is limited. The experimental results for classification-based object detection on canonical datasets including VOC 2007 (60.1\%), 2010 (56.4\%) and 2012 (56.3\%) show obvious improvement in mean average precision (mAP) with simple linear support vector machines.
\end{abstract}

\section{Introduction}

Object detection is a fundamental and crucial problem in computer vision. 
One of the most heavily studied paradigms and the most prominent example for object detection is deformable part-based models (DPM) algorithm~\cite{lsvm-pami}. It combines a set of discriminatively trained parts in a star model which is called pictorial structure~\cite{PictorialStructures1973,felzenszwalb2000efficient,felzenszwalb2005pictorial}. The part filters in DPM are based on hand-crafted Histogram of Gradients descriptors~\cite{dalal2005histograms}. However, the progress has been slow during 2010-2012 in the canonical visual recognition task PASCAL VOC object detection~\cite{everingham2010pascal} with hand-crafted visual features. 

In the last years, more and more works focus on Deep Convolutional Neural Networks (CNNs) and achieve great success. CNNs were firstly introduced in 1980 by Kunihiko Fukushima~\cite{cnns1980}, and Yann LeCun et al. improved it in 1998~\cite{lecun1998gradient}. This model was initially applied to handwritten digit recognition~\cite{handwritten1989} and OCR~\cite{lecun1998gradient}. Recently CNNs have been well applied into lots of visual recognition systems in a variety of domains. With the introduction of large labeled image databases~\cite{deng2009imagenet} and the massive parallel computations of GPU implementations, the large scale CNNs have become the most accurate method for generic object classification~\cite{krizhevsky2012imagenet} , detection~\cite{girshick14CVPR,sermanet-iclr-14,szegedy2013deep} and segmentation tasks~\cite{hariharan14sds,gupta14rcnndepth}. 

\begin{figure}[t]
	\begin{center}
		\includegraphics[width=0.5\textwidth]{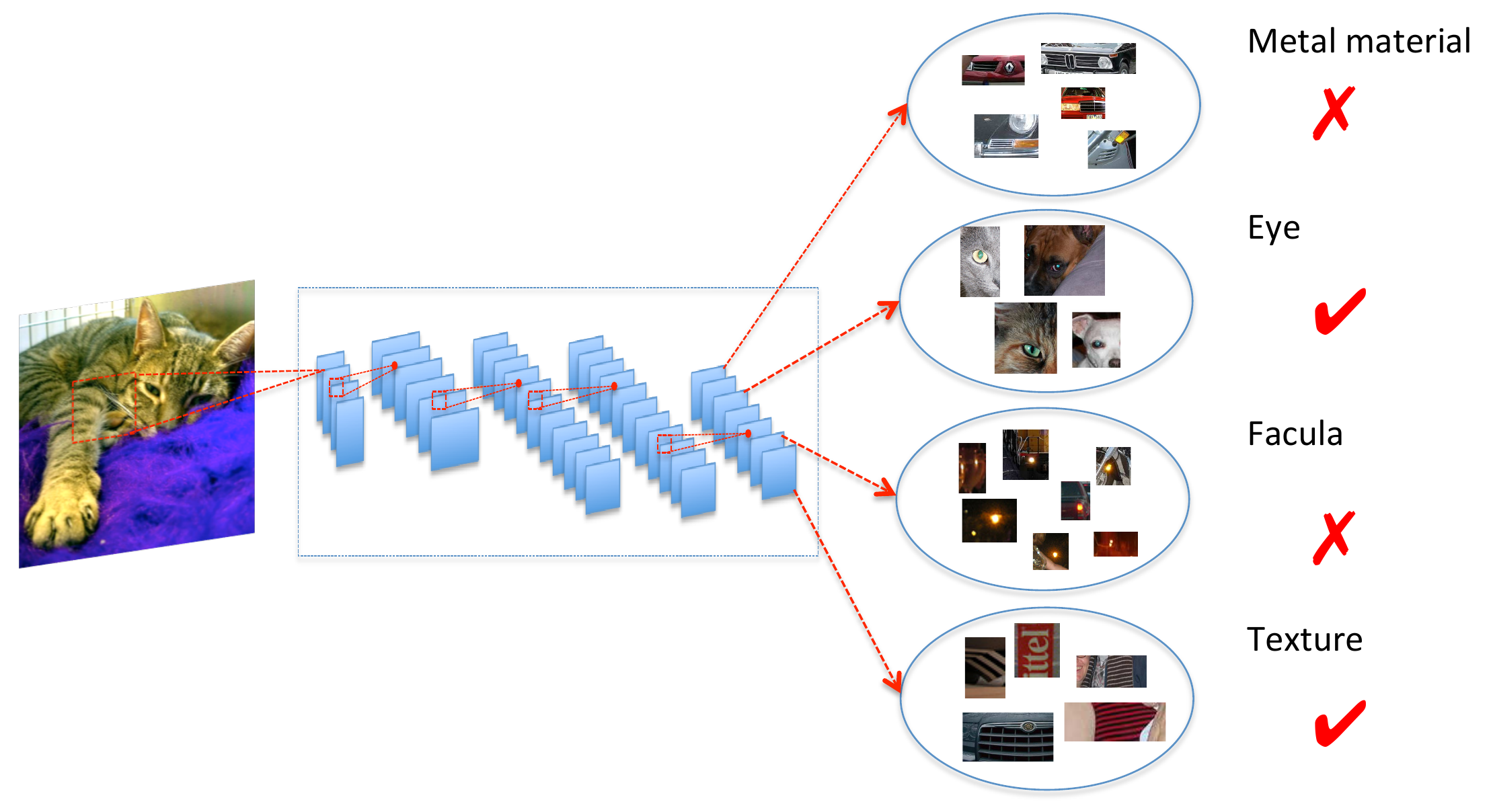}
	\end{center}
	\caption{Illustration showing the main motivation of this paper. For a parameter-trained network, correspondence between pool$_5$ features and concepts of each channel would also be fixed, such as shape, texture and material properties. It is clear that some concepts are helpful and some are unfriendly for classification. }
	\label{fig:concept}
\end{figure}

In 2012, Krizhevsky et al.~\cite{krizhevsky2012imagenet} designed a large CNN with 60 million parameters and 650,000 neurons and obtained substantial improvement of image classification accuracy on the ImageNet Large Scale Visual Recognition Challenge (ILSVRC) ~\cite{imagenet2012, deng2009imagenet}. In 2013, a joint deep learning architecture was proposed for pedestrian detection, which combines a CNN architecture with the DPM algorithm~\cite{ouyang2013joint}. Four components are contained in this framework: feature extraction, part deformation handling, occlusion handling, and classification. In 2014, Girshick et al.~\cite{girshick14CVPR} proposed a scalable detection algorithm called R-CNN and showed that R-CNN can obtain dramatically higher object detection performance on PASCAL VOC as compared to algorithms based on HOG-like features. R-CNN is a region-based algorithm that bridges the gap between image classification and object detection by operating within the ``detection with regions'' paradigm~\cite{Uijlings13} instead of the sliding-window strategy.  

For further improving detection performance, several methods were studied before. One kind approach is to manipulate the training images by different operations, such as rotation, scaling, shearing and stretching, and  then merge these transformed images into training set for training a more powerful detector which will improve the view-invariant representations~\cite{hui2013direct}. Another kind of approach is to perform local transformations in feature learning algorithms like Restricted Boltzmann Machine (RBM) and autoencoders, which combines various kinds of transformed weights (filters) and expands the feature representations with transformation-invariance~\cite{sohn2012learning}. In reality, occlusion and inconsistent property such as the same crowds or objects with different colors, texture or material properties often exists between training and testing data sets. So simply considering invariance is still far from enough. 

Considering this deficiency, we hold the opinion that establishing correspondence on middle-level features with input images is required. Some works have paid	attention to this idea~\cite{girshick14CVPR, long2014convnets}. Ross Girshick et al.~\cite{girshick14CVPR} visualize activation units at $5th$ convolutional layer in the middle of each channel after pooling operation (the pool$_5$ feature map is 6$\times$6$\times$256 dimensional) and find that units in pool$_5$  somewhat characterize concepts of objects (people or text), or texture and material properties, such as dot arrays and specular reflections. If we dropout some activation units to zeros, it seems that we perform some changes in the input images of CNN. If activation units with large intra-class and small inter-class variations set to zeros in the training phase, what would happen for object detection? Fortunately our experimental results give us positive answers. 

Inspired by this, we propose a feature edit algorithm by finding out the distribution of concepts that pool$_5$ feature units correspond to. Our method is an entropy-based model which is to compute the probability that each concept owns in training data and drop out some lowest ones. That is to say we drop out those units unfriendly to classification in the pool$_5$ feature maps of CNN. Our algorithm named as $Feature Edit$ is different from feature selection algorithms mainly for dimensionality reduction in order to lower the computational complexity and improve generalization~\cite{EigenstetterO12}.
 
Automatic feature selection can be formulated as the problem of finding the best subset $S$ of features from an initial, and maybe a very large set of features $F$($i.e., S\subset F$). Since the ultimate goal is to improve performance, one could define the optimal subset of features as which provides the best classification ability in the given task. We measure it by a criterion function $G$ as $c=G(S,D,M)$ where value $c$ denotes the model classification ability, $D$ denotes the data set used and $M$ denotes the model parameters applied in the task. Our proposed feature edit algorithm belongs to the so-called $filter$ methods which could be heuristically defined and the superiority would be evaluated by the learning algorithms independent of the classifiers. The criterion function of our algorithm can be formulated as $c = G(E,\,D,\,M) $ where $E$ is the edited version of $F$ (i.e. feature sets are extracted from all training images by CNN). We use $5th$ convolutional layer features after max pooling operations for editing but still map to $7th$ layer's feature maps by multiplying the weights of $w_6$ and $w_7$ and then feed the new feature set to the boosted linear-SVM for training.

This paper makes two important contributions: (1) We find out that we can easily obtain slightly modified CNN features by performing more dropouts in pool$_5$ without decreasing the ultimate object detection performance. (2) We present an entropy-based model to mark the subsets of pool$_5$ feature maps, which do not benefit for classification. We achieve obviously higher performance of 60.1\%, 56.4\% and 56.3\% mAP on PASCAL VOC 2007, VOC 2010 and VOC 2012, which are higher than R-CNN and other competition approaches.
\begin{figure}[t]
	\begin{center}
		\includegraphics[width=0.5\textwidth]{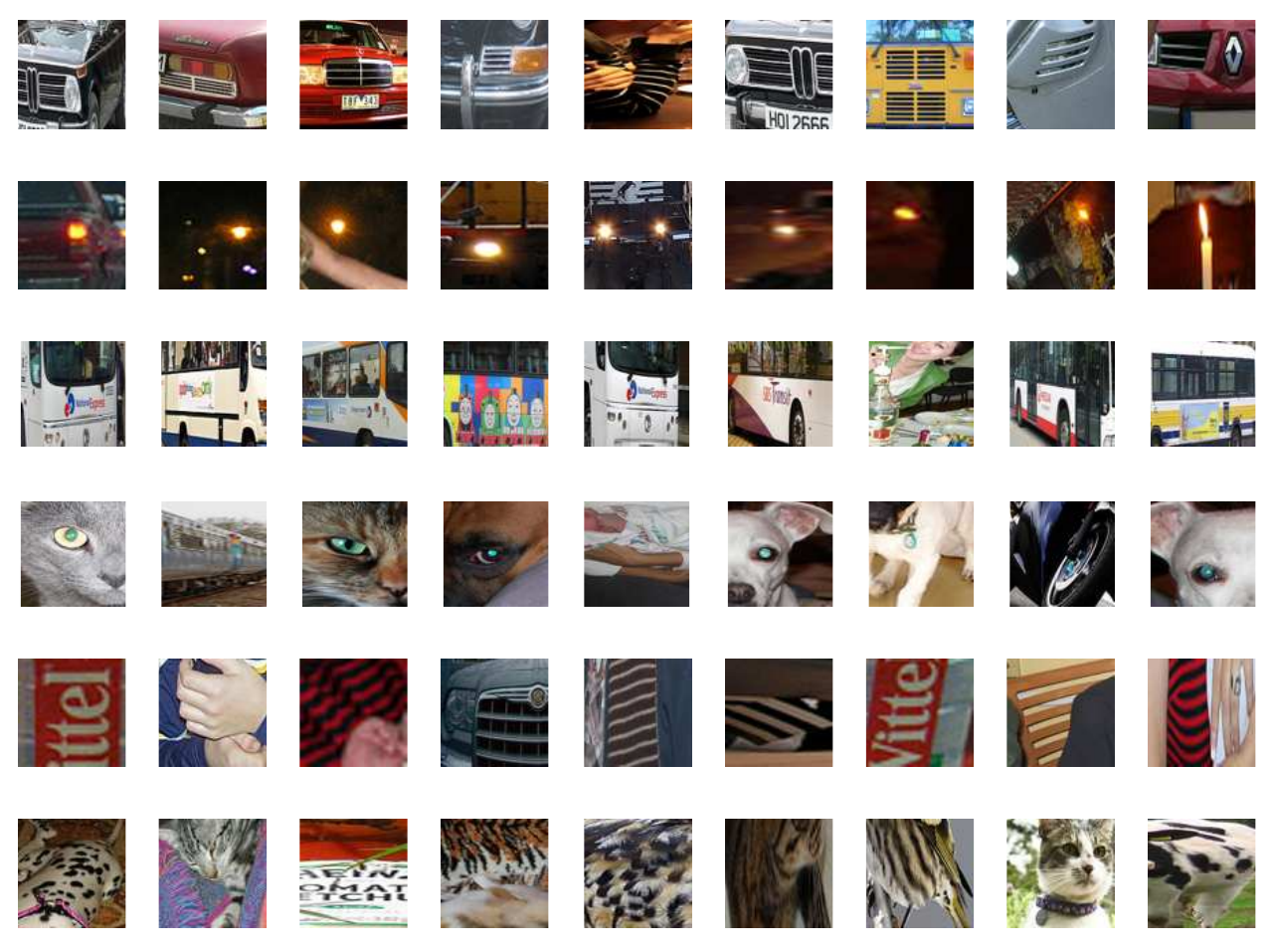}
	\end{center}
	\caption{Illustration showing top 9 regions (receptive field) for six pool$_5$ units by a trained network. Some units are aligned to shapes, such as bus(row 3) or eye (row 4). Other units capture texture and material properties, such as dot arrays (row 6) and specular reflections (row 2).}
	\label{fig:concept}
\end{figure}
\section{Concepts Correspondence}\label{sec2}
 Establishing correspondence on a finer level than object category is required to understand convolutional neural networks. Jonathan Long et al.~\cite{long2014convnets} studied the effectiveness of activation features for tasks requiring correspondence and present evidence that CNN features localize at a much finer scale than their receptive field sizes.
 On the other hand, Zeiler and Fergus~\cite{zeiler2014visualizing} introduced a novel visualization technique that gives insight into the function of intermediate feature layers, which could explain the impressive classification performance of the large convolutional neural networks. Each layer in classification CNN models shows the hierarchical nature of the features in the network. Here we focus on the detection networks which are slightly different from classification ones. Detection networks are fine-tuned from neat objects, but classification networks are trained on global images with large background area. In this paper, we just concentrate on the $5th$ layer features and how CNN models identify variant of objects.

Our concepts correspondence method follows the feature visualization in R-CNN~\cite{girshick14CVPR} which sorts massive region proposals from highest to lowest activation. We make several slight modification in our experiments, one is the activation units index, the pool$_5$ feature map is 6 $\times$ 6 $\times$ 256 dimensional (xIndex$\times$yIndex$\times$channel), we move the activation indexes from $(3,3,channel)$ to $Max (3:4,3:4,channel)$. Another modification is the region proposals, proposals in the original method are obtained by selective search. However, our proposals are randomly sheared from every object area. In each object, we shear about one thousand patches to order and perform non-maximum suppression for filtering.

Each row in Figure ~\ref{fig:concept} shows the top 9 activations for a pool$_5$ unit from the CNN that we fine-tuned on VOC 2012 trainval. It implies that these units are aligned to shapes, textures, properties and other concepts. So if we set some particular channel units to zeros, the input image will be manipulated indirectly on concept-level.

\section{Maximum Entropy Model}
\subsection{Preliminaries}
Let $\mathcal{X} \subseteq  \mathbb{R}^d$ and $\mathcal{Y}$ be the input and output space\footnote {Input space means the pool$_5$ feature map; Output space means the measured concept presentation for each channel.}, $X \in \mathcal{X}$ and  $Y \in \mathcal{Y}$. We define a train sample as $(x_i,y_i),i=1,2,\dots,256$. 

The correspondence between each channel feature vectors and concept space is defined by $kurtosis$ which measures peakedness of a distribution. 	
$kurtosis$ is defined as
\[x(a) = \frac{{E[o^4]}}{{{E^2}[o^2]}} - 3\] where ${o^\alpha} = {({a_i} - \bar a)^\alpha}$ and $E[\cdot]$ is the expectation operator for vector segments. $a_i$ denotes the units in $ith$ channel of pool$_5$ feature map. 
The -3 operator is to let normal-distribution kurtosis approach zero.

Let $\{f_t(x,y),t=1,2,\dots,T\}$ be the feature function for different classes, which is a real-valued function. $T$ denotes the category number in training dataset. ${\tilde p}(x,y)$ denotes the empirical distribution on training data. Feature function is defined as 
\[{f_t}(x,y) = \left\{ {\begin{array}{*{20}{c}}
1&{if\;\tilde p(x,y) > threshold}\\
0&{else}
\end{array}} \right.\]
The expectation of empirical distribution ${\tilde p}(x,y)$ is defined as
\[{E_{\tilde p}}(f_t) = \sum\limits_{x,y} {\tilde p(x,y)f_t(x,y)} \]
The expectation of distribution $p(x,y)$ is defined as
\[{E_{{p}}}(f) = \sum\limits_{x,y} {\tilde p(x)p(y|x)f_t(x,y)} \]

\subsection{Problem Definition}

Maximum entropy~\cite{berger1996maximum} can be used to estimate any probability distribution in its general formulation. In this paper we are interested in improving the performance to classify different objects. Thus we limit our further discussion to learn conditional distributions from labeled training data. 
The motivation of our idea is to find out the probability distribution that concepts are representative of the input object images and dropout those unfriendly ones for final classification.

Suppose that we are given $T$ feature functions $f_t$ for each class in training data. We would like our model to accord with the statistics that we train.
Our goal is to extract a set of facts about the decision-making process from the training data that will aid us in constructing a model of this process. So we would like $p$ to lie in the subset $\mathcal{C}$ of $\mathcal{P}$ which defined by

\[\mathcal{C} \equiv \left\{ {p \in \mathcal{P}|E({f_t}) = \tilde E({f_t})\quad for\;t \in \{ 1,2, \cdots ,T\} } \right\}\]

The conditional entropy is defined as 
\[H(p) =  - \sum\limits_{x,y} {\tilde p(x)p(y|x)\ln p(y|x)} \]

In maximum entropy, we use the training data to set constraints on the conditional distribution.
To select a model from a set $\mathcal{C}$ of allowed probability distribution, choose the model $p^* \in \mathcal{C}$ with maximum entropy $H(p)$
\[{p^ * } = \mathop {\arg \max }\limits_{p \in\mathcal{C}} \,H(p)\]

\begin{figure*}[t]
	\begin{center}
		\includegraphics[width=1\textwidth]{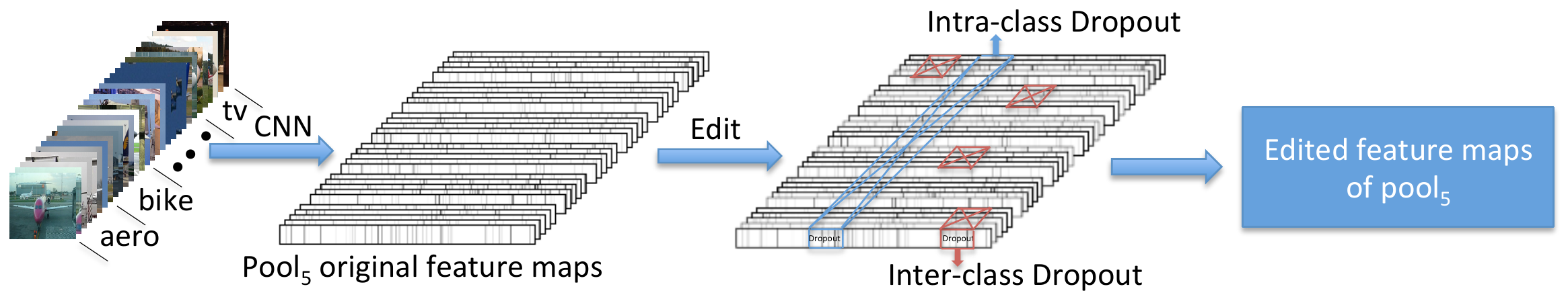}
	\end{center}
	\caption{Illustration showing the edit procedure. From left to right: (1) Input regions (ground truth), which are resized to 227$\times$227, (2) Pool$_5$ feature map extracted from CNN, pool$_5$ means the $5th$ convolutional layer features after max pooling operation, (3) Edit operation, (4) Outputs of the edited features.}
	\label{fig:system}
\end{figure*}

\section{Nonparametric  Algorithm for $p^*$}
Instead of referring to original constrained optimization problem as the $primal$ problem, we propose a nonparametric method to find out $p^*$. We call our method as $Edit$ algorithm which is based on the intra-class and inter-class variations of activation value $x_i$ in each channel.

\subsection{Algorithm Details}

\begin{itemize}
	\item Intra-class Edit 
	
	In this stage we would like to find out the subset of each channel features which has the largest intra-class variations.
	
	We define the training set $\{x_i\}$ as ${K_{ji}^C({\cdot})}$ which denotes the value of $Cth$ class in $jth$ training sample and $ith$ channel segment, $i=1,2,\cdots,256$, $j=1,2,\cdots,N_C$,  $C=1,2,\cdots,T$, $N_C$ is the number of training examples in each class $C$.
	
	Then compute the variance of each feature segment's statistic in class $C$ which is defined as
	\[V_i^C = \frac{1}{N_C}\sum\limits_{j = 1}^{N_C} {{{(K_{ji}^C({a^i}) - \bar K_{ji}^C({a^i}))}^2}} \]
	where $V_{i}^{C}$ denotes the variance of $Cth$ class training feature vectors in $ith$ channel feature segment. The larger of $V_{i}^C$ means channel $i$ is unsteady and unfriendly for classification.
	
	Following it we compute intra-class $p^*$ by $V_i^C$
	
	\[{p_i}^*(intra) = \frac{{{V_i}^C}}{{\sum\limits_i {{V_i}^C} }}\]

	\item Inter-class Edit 
	
	We find out subsets with the smallest variations for all classes. First compute the mean value of the statistics at channel $i$ in $Cth$ class:
	\[\bar K_i^C(a^i) = \frac{1}{N_C}\sum\limits_{j = 1}^{N_C} {K_{ji}^C}(a^i) \]
	
	Then compute the mean value of the statistics in all $T$ classes:
	\[\bar K_i^A(a^i) = \frac{1}{T}\sum\limits_{C = 1}^T {\bar K_{ji}^C}(a^i) \]
	where $\bar K_i^A(\cdot)$ denotes the average statistic of all classes.
	
	The variance of the statistics in all classes is defined as
	\[V_i^A = \frac{1}{T}{\sum\limits_{C = 1}^T {(\bar K_i^C({a^i}) - \bar K_i^A({a^i}))} ^2}\]

     Then we compute inter-class $p^*$ 
	\[{p_i}^*(inter) = \frac{{{V_i}^A}}{{\sum\limits_i {{V_i}^A} }}\]
\end{itemize}	

	For the original training feature set ${F_{CNN}} = \{ F_1, F_2,\cdots, F_{N_C}\} \in {\Re ^{N_C\times k}}$, where $k$ is the length of one CNN feature vector. $F_i( i=1,2,\cdots,N_C)$ denotes one feature vector extracted from an image.
The edited feature values are defined as	
	\[ x_{edit}=x \otimes f(x,y) \]
where $f(x,y)$ is defined by $p^*$ with 20\% intra-class and 30\% inter-class $threshold$. And then we can obtain $F_{Edit}$ by $x_{edit}$. For each channel units in $F_{CNN}$, if $x_i=0$, dropout all units in this channel to zeros. Pipeline is shown in Figure \ref{fig:system}. 

	Up to now, we have obtained edited features $F_{Edit}$ by dropping out selected channels in $pool_5$ feature maps according to the proposed algorithm. Because of the same dimensionality with original pool$_5$ feature maps $F_{Ori}$, $F_{Edit}$ can feed directly to the fully-connected layers of fc$_6$ and fc$_7$.

\begin{table*}\small
	\begin{center}
		\begin{tabular}{l|l|l}
			\hline
			VOC 2007 & aero \ bike \  bird \ boat \ bottle \ bus \ \ car \ \ cat \ chair \ cow \ table \ dog \ horse \ mbike \ person \ plant \ sheep \ sofa \ train \ \ tv & mAP\\
			\hline
			FE\_R &  69.1\ \ 72.7\ \ 54.9\ 42.0\ \ \ 35.0\ \ \ 66.1\ 74.1\ 64.9\ 36.8\ \ 67.7\ \ 55.1\ \ 63.6\ \ \ 67.3\ \ \ \ \textbf{71.9}\ \ \ \ \ 57.7\ \ \ \ 32.0\ \ \ \ 65.5\ \ \ 51.6\ \ 64.1\ 64.9  &58.8  \\
			FE\_S &  68.0\ \ 69.1\ \ 52.9\ 41.5\ \ \ 35.1\ \ \ 66.1\ 72.9\ 67.4\ 36.2\ \ 65.4\ \ 51.8\ \ 61.0\ \ \ 65.8\ \ \ \ 70.9\ \ \ \ \ 55.7\ \ \ \ 31.7\ \ \ \ 64.7\ \ \ 51.9\ \ 61.5\ 64.2  &57.7  \\
			FE\_E &  \textbf{71.3}\ \ 71.6\ \ 56.1\ 42.2\ \ \ 37.1\ \ \ 67.2\ 74.4\ 67.9\ 36.6\ \ 68.2\ \ 54.3\ \ 64.7\ \ \ 70.2\ \ \ \ 70.9\ \ \ \ \ 58.9\ \ \ \ 34.7\ \ \ \ \textbf{66.3}\ \ \ \textbf{53.8}\ \ 64.2\ 67.3  &59.8  \\
			FE\_M & 71.0\ \ 72.1\ \ 55.2\ 41.5\ \ \ 36.3\ \ \ \textbf{69.5}\ \textbf{74.7}\ 67.2\ \textbf{37.2}\ \ \textbf{68.6}\ \ 57.1\ \ \textbf{64.7}\ \ \ 69.8\ \ \ \ 71.8\ \ \ \ \ \textbf{59.1}\ \ \ \ \textbf{35.3}\ \ \ \ 65.8\ \ \ 52.9\ \ 64.4\ 67.8  &\textbf{60.1}  \\
			\hline
			R-CNN & 68.1\ \ \textbf{72.8}\ \ 56.8\ \textbf{43.0}\ \ \ 36.8\ \ \ 66.3\ 74.2\ 67.6\ 34.4\ \ 63.5\ \ 54.5\ \ 61.2\ \ \ 69.1\ \ \ \ 68.6\ \ \ \ \ 58.7\ \ \ \ 33.4\ \ \ \ 62.9\ \ \ 51.1\ \ 62.5\ 64.8  &58.5  \\
			SPP & 68.6\ \ 69.7\ \ \textbf{57.1}\ 41.2\ \ \ \textbf{40.5}\ \ \ 66.3\ 71.3\ \textbf{72.5}\ 34.4\ \ 67.3\ \ \textbf{61.7}\ \ 63.1\ \ \ \textbf{71.0}\ \ \ \ 69.8\ \ \ \ \ 57.6\ \ \ \ 29.7\ \ \ \ 59.0\ \ \ 50.2\ \ \textbf{65.2}\ \textbf{68.0}  &59.2  \\
			DNP+R &  \qquad \qquad \qquad \qquad \qquad \qquad \qquad \qquad \qquad \qquad \qquad  ---  & 46.1  \\
			Regionlets &  54.2\ \ 52.0\ \ 20.3\ 24.0\ \ \ 20.1\ \ \ 55.5\ 68.7\ 42.6\ 19.2\ \ 44.2\ \ 49.1\ \ 26.6\ \ \ 57.0\ \ \ \ 54.5\ \ \ \ \ 43.4\ \ \ \ 16.4\ \ \ \ 36.6\ \ \ 37.7\ \ 59.4\ 52.3  &41.7  \\
			
			Szegedy &29.2\ \ 35.2\ \ 19.4\ 16.7\ \ \ \ \ 3.7\ \ \ 53.2\ 50.2\ 27.2\ 10.2\ \ 34.8\ \ 30.2\ \ 28.2\ \ \ 46.6\ \ \ \ 41.7\ \ \ \ \ 26.2\ \ \ \ 10.3\ \ \ \ 32.8\ \ \ 26.8\ \ 39.8\ 47.0  &30.5  \\
			DPM v5  & 33.2\ \ 60.3\ \ 10.2\ 16.1\ \ \ 27.3\ \ \ 54.3\ 58.2\ 23.0\ 20.0\ \ 24.1\ \ 26.7\ \ 12.7\ \ \ 58.1\ \ \ \ 48.2\ \ \ \ \ 43.2\ \ \ \ 12.0\ \ \ \ 21.1\ \ \ 36.1\ \ 46.0\ 43.5  &33.7  \\
			\hline
			$\bigtriangleup_{R-CNN}$   & +2.9\ \ -0.7\ \ \ -0.4\ \ -1.5\ \ \ \ -0.5\ \ +3.2\ +0.5\ \ -0.4\ +2.8\ +5.1\ \ +2.6\ \ +3.5\ \ \ +0.7\ \ \ +3.2\ \ \ \ \ +0.4\ \ \ \  +1.9\ \ \ \ +2.9\ \  +1.8\ \ +1.9\ +3.0  &+1.6  \\
			\hline
		\end{tabular}
	\end{center}
	\caption{Detection average precision(\%) on PASCAL VOC 2007 test. Rows 1-4 show our experimental results. Feat\_R: Random feature edit algorithm; Feat\_S: Images shearing algorithm; Feat\_E: Only using edited features for training; Feat\_M: Merging the original features and edited features for training. Rows 5-10 show other competition approaches. (R-CNN~\cite{girshick14CVPR}; SPP (without combination)~\cite{he2014spatial}; DNP+R~\cite{Zou:DNPRegionlets14}; Regionlets~\cite{wang2013regionlets}; Szegedy et al.~\cite{szegedy2013deep}; DPM v5~\cite{lsvm-pami}). Row 11 shows the differences between FE\_M and R-CNN.}
	\label{2007results}
\end{table*}

\section{Experiments and Discussions}
We evaluate performance on the datasets: Pascal Visual Object Challenge (VOC) 2007~\cite{pascal-voc-2007} and VOC 2010-2012~\cite{pascal-voc-2010}. VOC 2007 dataset contains 5011 trainval (training + validation) images and 4952 test images over 20 classes. 10103 trainval images and 11635 test images are in VOC 2010. The data sets are identical for VOC 2011 and 2012 which both contain 11540 images in trainval set and 10991 images in test set. 

\subsection{Training Stage}
There are two parts in training stage: train parameters of convolutional neural network and train a linear classifier. Details are as follows.

\subsubsection{Network Parameters Training}

We use the Caffe~\cite{Jia13caffe} implementation of the CNN defined by Krizhevsky et al.~\cite{krizhevsky2012imagenet}, which is used in various kinds of domains such as fine-grained category detection~\cite{zhang14finegrained} and object detection~\cite{girshick14CVPR}. It consists of total seven layers, the first five are convolutional and the last two are fully connected. Our training strategy is supervised pre-training on a large auxiliary dataset (imagenet 2012 trainval dataset) and domain-specific fine-tuning on VOC 2012 trainval dataset. This strategy is popular in recent years such as R-CNN~\cite{girshick14CVPR}. Although VOC 2012 is the largest in PASCAL VOC datasets, it is still not enough to fine-tune the parameters to a good location. Thanks to the improvement of generic objectness measures which produce a small set of candidate object windows from raw images, we select all candidate object proposals with $\ge$ 0.6 IoU overlap with a ground truth box to rich the fine-tuning dataset. Two existing high performance objectness detectors have been tried: BING~\cite{BingObj2014} and selective search~\cite{Uijlings13}. We find that BING is faster than selective search but the IoU overlap with ground truth is lower than the latter. In this paper we use selective search for pre-detecting, but if you care efficiency more, BING will be a better choice. CNN fine-tuning is run for 70k SGD iteration on VOC 2012 trainval dataset and selected windows. The CNN we used requires a fixed-size input of 227$\times$227 pixels. R-CNN~\cite{girshick14CVPR} has evaluated two approaches for transforming object proposals into CNN inputs and finds that warping with context padding ($p$=16 pixels) outperformed other approaches (more details in R-CNN). Finally a fixed-dimensional feature vector from each proposal is extracted. Our CNN parameters and algorithm codes will be released soon.

\subsubsection{Linear-SVM and Regression}
A linear classifier is trained on different feature sets respectively, including random edited features, image shearing features, two stages edited features and merging edited features with original features in our experiments (more details in Table~\ref{2007results}). $L2$ regularization and $L1$ hinge loss are used as the loss function.

Inspired by the bounding-box regression employed in DPM~\cite{lsvm-pami} and R-CNN~\cite{girshick14CVPR}, we also train a linear regression model to predict a new detection window using the $5th$ layer features after pooling operation for selective search region proposals, which is to reduce the localization errors.

\subsection{Testing Stage}

At testing stage, the sliding-window approach and objectness approach are both considered. Although the developments of GPU and parallel technology are rapidly moving forward, computational cost of extracting features from large scale CNNs is still expensive. Precise localization within a sliding-window paradigm is not suitable for combining with large scale CNNs. Thanks to the development of objectness, which has been successful for object detection~\cite{BingObj2014,Uijlings13}, we apply  selective search~\cite{Uijlings13} with ``fast'' model to generate about 2000 category-independent region proposals for an input image at the test phase like R-CNN~\cite{girshick14CVPR} and the object detection task is transformed to a standard image classification task. Then non-maximum suppression with 30\% threshold is used on the scored windows.

\begin{table*}\small
	\begin{center}
		\begin{tabular}{l|l|l}
			\hline
			VOC 2010 & aero \ bike \  bird \ boat \ bottle \ bus \ \ car \ \ cat \ chair \ cow \ table \ dog \ horse \ mbike \ person \ plant \ sheep \ sofa \ train \ \ tv & mAP\\
			\hline
			FE\_M & \textbf{74.8}\ \ \textbf{69.2}\ \ \textbf{55.7}\ \textbf{41.9}\ \ \ \textbf{36.1}\ \ \ \textbf{64.7}\ \textbf{62.3}\ 69.5\ \textbf{31.3}\ \ \textbf{53.3}\ \ \textbf{43.7}\ \ 69.9\ \ \ \textbf{64.0}\ \ \ \ \textbf{71.8}\ \ \ \ \ \textbf{60.5}\ \ \ \ \textbf{32.7}\ \ \ \ \textbf{63.0}\ \ \ \textbf{44.1}\ \ \textbf{63.6}\ \textbf{56.6}  &\textbf{56.4}  \\
			\hline
			R-CNN &71.8\ \ 65.8\ \ 53.0\ 36.8\ \ \ 35.9\ \ \ 59.7\ 60.0\ \textbf{69.9}\ 27.9\ \ 50.6\ \ 41.4\ \ \textbf{70.0}\ \ \ 62.0\ \ \ \ 69.0\ \ \ \ \ 58.1\ \ \ \ 29.5\ \ \ \ 59.4\ \ \ 39.3\ \ 61.2\ 52.4  &53.7  \\
			Regionlets & 65.0\ \ 48.9\ \ 25.9\ 24.6\ \ \ 24.5\ \ \ 56.1\ 54.5\ 51.2\ 17.0\ \ 28.9\ \ 30.2\ \ 35.8\ \ \ 40.2\ \ \ \ 55.7\ \ \ \ \ 43.5\ \ \ \ 14.3\ \ \ \ 43.9\ \ \ 32.6\ \ 54.0\ 45.9  &39.7  \\
			SegDPM$^\dag$ & 61.4\ \ 53.4\ \ 25.6\ 25.2\ \ \ 35.5\ \ \ 51.7\ 50.6\ 50.8\ 19.3\ \ 33.8\ \ 26.8\ \ 40.4\ \ \ 48.3\ \ \ \ 54.4\ \ \ \ \ 47.1\ \ \ \ 14.8\ \ \ \ 38.7\ \ \ 35.0\ \ 52.8\ 43.1  &40.4  \\
			DPM v5$^\dag$   & 49.2\ \ 53.8\ \ 13.1\ 15.3\ \ \ 35.5\ \ \ 53.4\ 49.7\ 27.0\ 17.2\ \ 28.8\ \ 14.7\ \ 17.8\ \ \ 46.4\ \ \ \ 51.2\ \ \ \ \ 47.7\ \ \ \ 10.8\ \ \ \ 34.2\ \ \ 20.7\ \ 43.8\ 38.3  &33.4  \\
			\hline
			$\bigtriangleup_{R-CNN}$   &  +3.0\ \ +3.4\  +2.7\ +5.1\ \ \ +0.2\ \  +5.0\ +2.3\ \ -0.4\ +3.4\ \ +2.7\ \ +2.3\ \ -0.1\ \ \ +2.0\ \ \ \ +2.8\ \ \ \  +2.4\ \ \ \ +3.2\ \ \ \ +2.6\ \ \ +4.8\  +2.4\ +4.2  &+2.7  \\
			\hline
		\end{tabular}
	\end{center}
	\caption{Detection average precision(\%) on PASCAL VOC 2010 test. Row 1 shows our experimental results. Row 2-5 show other competition approaches. SegDPM$^\dag$~\cite{fidler2013bottom} and DPM v5$^\dag$~\cite{lsvm-pami} use context rescoring. Row 6 shows the differences between FE\_M and R-CNN.}
	\label{2010results}
\end{table*}

\begin{table*}\small
	\begin{center}
		\begin{tabular}{l|l|l}
			\hline
			VOC 2012 & aero \ bike \  bird \ boat \ bottle \ bus \ \ car \ \ cat \ chair \ cow \ table \ dog \ horse \ mbike \ person \ plant \ sheep \ sofa \ train \ \ tv & mAP\\
			\hline
			FE\_M & \textbf{74.6}\ \ \textbf{69.1}\ \ \textbf{54.4}\ \textbf{39.1}\ \ \ \textbf{33.1}\ \ \ \textbf{65.2}\ \textbf{62.7}\ 69.7\ \textbf{30.8}\ \ \textbf{56.0}\ \ \textbf{44.6}\ \ \textbf{70.0}\ \ \ \textbf{64.4}\ \ \ \ \textbf{71.1}\ \ \ \ \ \textbf{60.2}\ \ \ \ \textbf{33.3}\ \ \ \ \textbf{61.3}\ \ \ \textbf{46.4}\ \ \textbf{61.7}\ \textbf{57.8}  &\textbf{56.3}  \\
			\hline
			R-CNN &  71.8\ \ 65.8\ \ 52.0\ 34.1\ \ \ 32.6\ \ \ 59.6\ 60.0\ \textbf{69.8}\ 27.6\ \ 52.0\ \ 41.7\ \ 69.6\ \ \ 61.3\ \ \ \ 68.3\ \ \ \ \ 57.8\ \ \ \ 29.6\ \ \ \ 57.8\ \ \ 40.9\ \ 59.3\ 54.1  &53.3  \\
			\hline
			$\bigtriangleup_{R-CNN}$   &  +2.8\ \ +3.3\  +2.4\ +5.0\ \ \ +0.5\ \  +5.6\ +2.7\ \ -0.1\ +3.2\ \ +4.0\  +2.9\ \ +0.4\ \ \ +3.1\ \ \ \ +2.8\ \ \ \  +2.4\ \ \ \ +3.7\ \ \ \ +3.5\ \ \ +5.5\  +2.4\ +3.7  &+3.0  \\
			\hline
		\end{tabular}
	\end{center}
	\caption{Detection average precision(\%) on PASCAL VOC 2012 test. Row 1 shows our experimental results. Row 2 shows R-CNN algorithm results. Row 3 shows the differences between them.}
	\label{2012results}
\end{table*}

\begin{figure}[t]
	\begin{center}
		\includegraphics[width=0.5\textwidth]{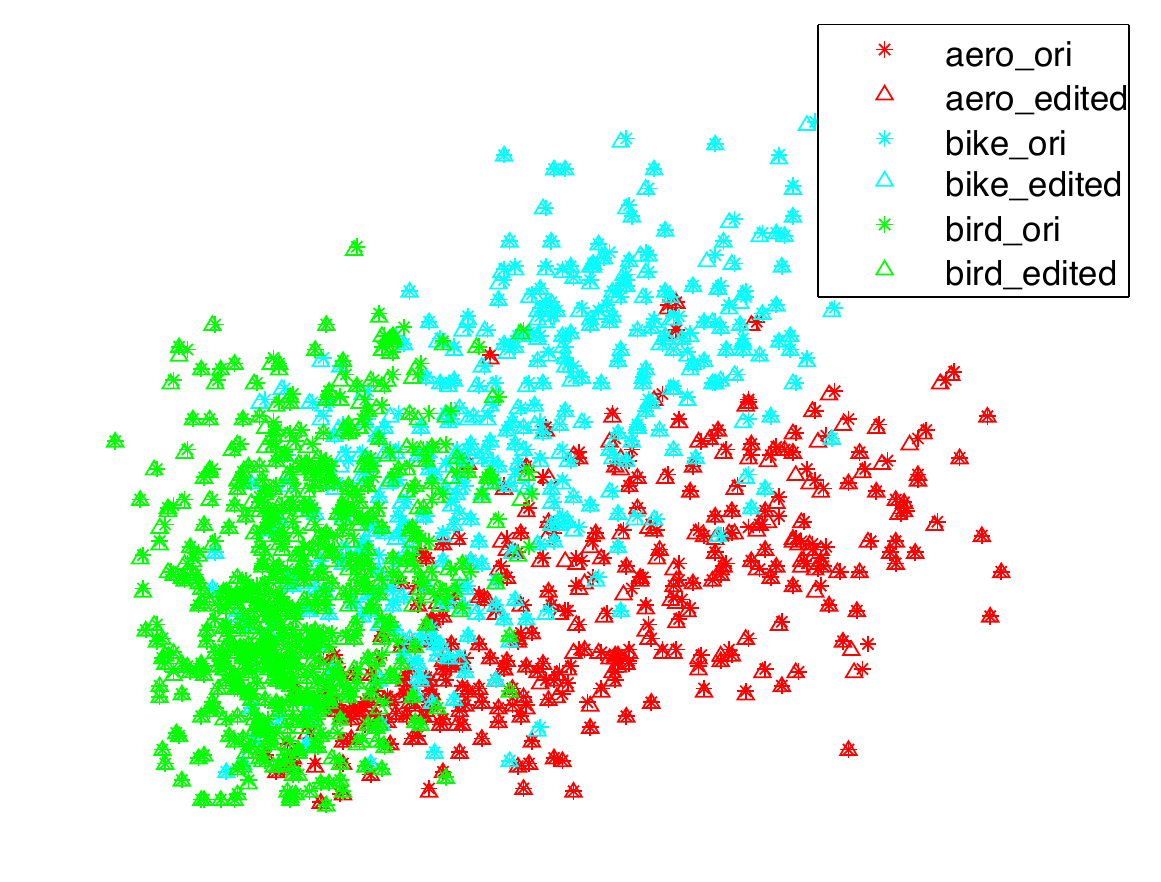}
	\end{center}
	\caption{Illustration showing the $5th$ layer CNN feature visualization of our system. The original features and edited features are visualized.}
	\label{FeatureVisual}
\end{figure}

\subsection{Feature Visualization}
Figure~\ref{FeatureVisual} shows the original and edited CNN features extracted from three categories of VOC 2007 trainval set using the $5th$ convolutional layer after pooling operation. In order to display the distribution of high-dimensional CNN features, we apply the principal component analysis (PCA) algorithm~\cite{wold1987principal} to reduce dimensionality and retain two components for visualization, which is inspired by~\cite{HinSal06}. From Figure~\ref{FeatureVisual} we can see that our edited features maintain the distribution of the original features and let the optimal hyperplane more refined which makes the boosted classifier more easy to classify different categories.

\subsection{Exp. ${\rm I}$: Comparison with Random Edit Algorithm}
We train our linear-SVM using train and validation set of VOC 2007. We compare our algorithm with random edit algorithm and the complete evaluation on VOC 2007 test is given in Table ~\ref{2007results}. 

\subsubsection{Random Edit Algorithm}
${R_i} = {F_i} \otimes W$, $i = 1,2, \cdots N_C$, where $\otimes$ denotes dot product, $W$ is a random binary vector with 0 or 1 and the length is $k$. $\frac{{m(0)}}{{m(1)}} = {\rm{threshold}}$, and $m(\cdot)$ denotes the number of $(\cdot)$ in the random vector $W$. And $R_{Edit}=\{R_1,R_2,\cdots,R_{N_C}\}$. 

We compare the results of random feature edit with our edit algorithm. Results are shown in Table~\ref{2007results}. 
We find that randomly editing CNN feature units  can also obtain competitive performance. This phenomenon proves that our CNN feature edit algorithm is efficient and powerful. Although performance of random feature edit is slightly lower compared to our two stages editing algorithm. The reasons are clear and can be explained between the training set and testing set. The deviation of properties between these two sets do not exist with the same ratio, so editing with propensities is better than random operation.

\begin{figure*}
	\centering
	\subfigure[cow]{\includegraphics[width=0.24\textwidth]{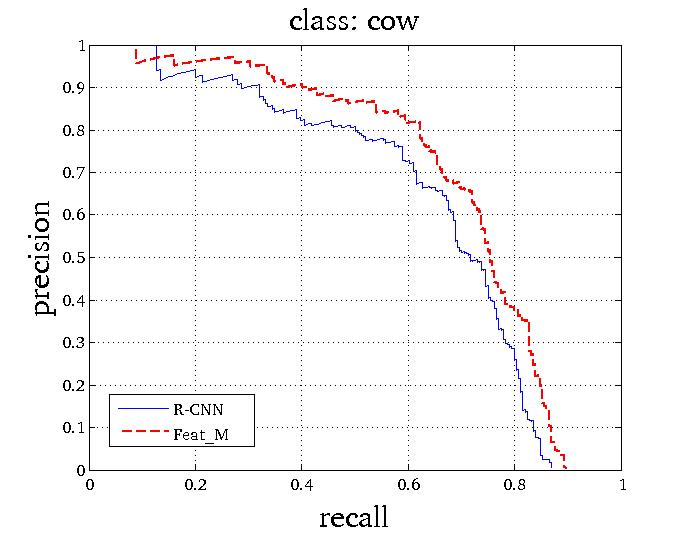}}
	\subfigure[dog]{\includegraphics[width=0.24\textwidth]{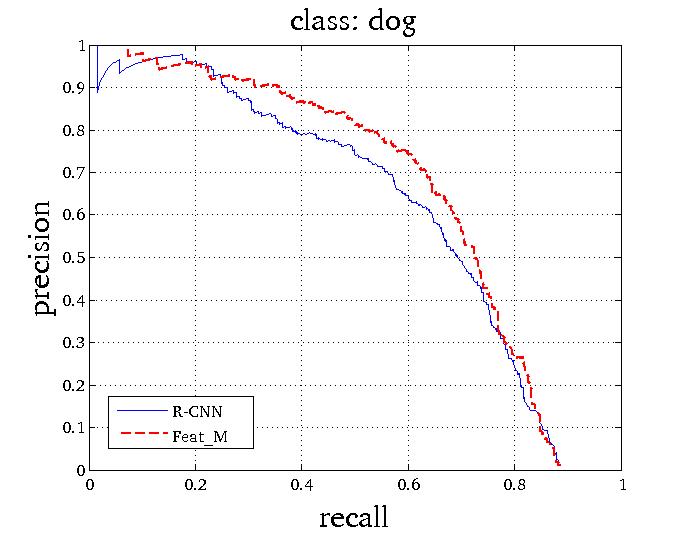}}
	\subfigure[boat]{\includegraphics[width=0.24\textwidth]{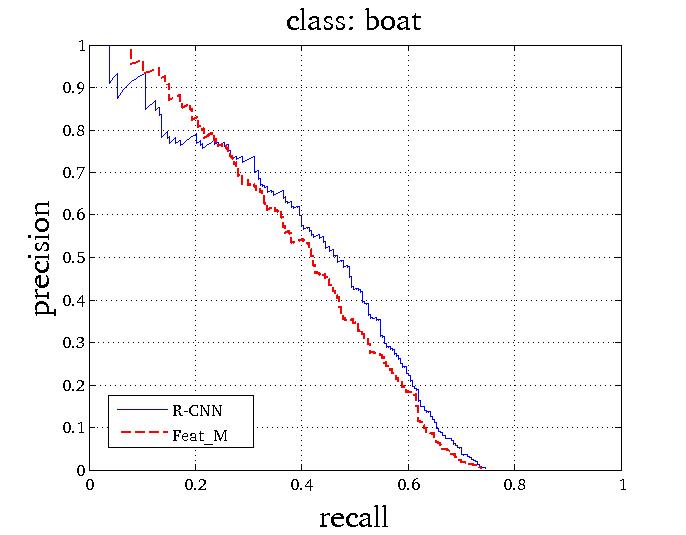}}
	\subfigure[bottle]{\includegraphics[width=0.24\textwidth]{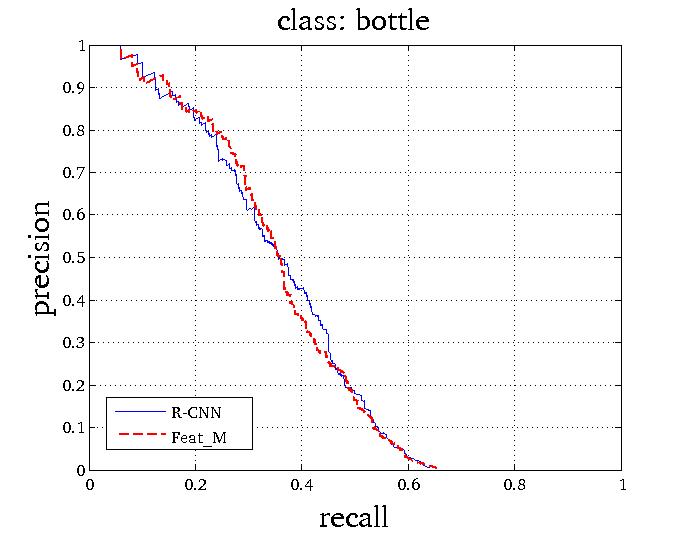}}
	\caption{Four representative categories of Precision/Recall curves on VOC 2007 for R-CNN results compared with Feat\_M results. (a) and (b): The categories which our performance is better than R-CNN; (c) and (d): The categories which our performance has a little slight decline compared with R-CNN.}
	\label{PRcurve}
\end{figure*}

\begin{figure}
	\centering
	\subfigure[Drops in $Boat$]{\includegraphics[width=0.45\textwidth]{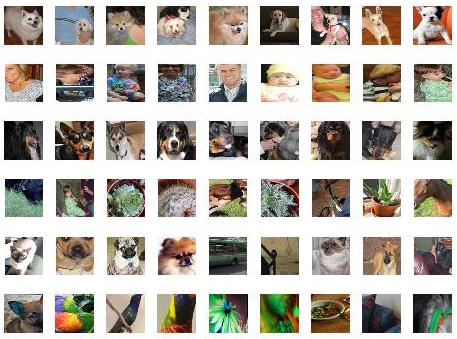}}
	\subfigure[Drops in $Person$]{\includegraphics[width=0.45\textwidth]{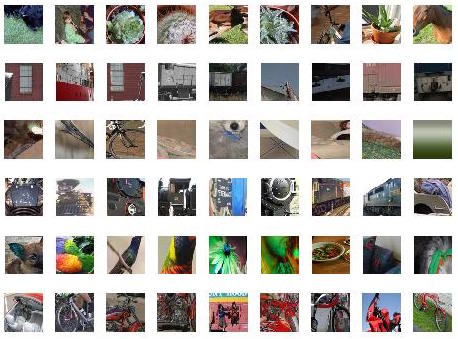}}
	\caption{Illustration showing the abandoned channels of $Boat$ and $Person$. These dropped features are helpless for describing their own categories. It's interesting that row 4 and 6 in class $Boat$ also appear in class $Person$ of row 1 and 5.}
	\label{drop}
\end{figure}

\subsection{Exp. ${\rm II}$: Comparison with Image Shearing}
We evaluate the approach that enriches training image set by merging sheared images (man-made object parts) and original ones whether can obtain the similar improvement if compared to our editing algorithm. We randomly shear one sub-image from an original object image in the training dataset, and then merge these sheared images into training set. The complete results are given in Table~\ref{2007results}. From the results, we can see that merging sheared images is harmful to the whole system. We consider the reasons are as follows: (1) The sub-images are randomly sheared without priori information, if some of these images have no discrimination between them, such as the background, people's chest or the small parts of the sofa. These sub-images will harm the performance of the boosted classifier. (2) The CNN feature units do not simply represent the parts of objects, but also characterize the concepts and properties such as colors, shapes and materials. Editing CNN features is more meaningful than image shearing. Shearing operation could not handle the separation of properties.

\subsection{Exp. ${\rm III}$: Merge or Not}
Simply using edited features and merging with original features for training are both evaluated in our experiments. The performance of  merging strategy is improved to 60.1\%.

\begin{figure*}
	\centering
	\subfigure[R-CNN]{\includegraphics[width=0.49\textwidth]{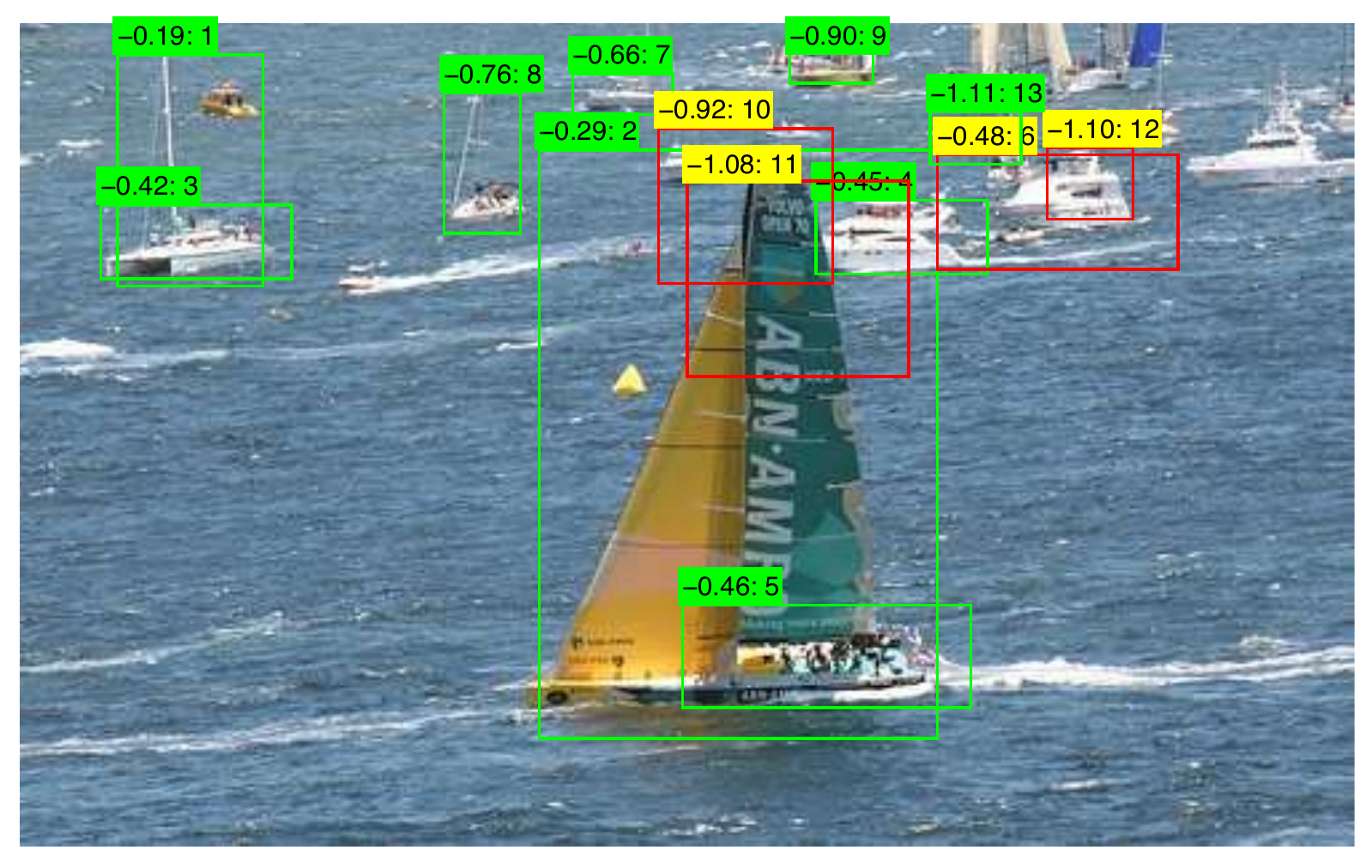}}
	\subfigure[FE\_M]{\includegraphics[width=0.49\textwidth]{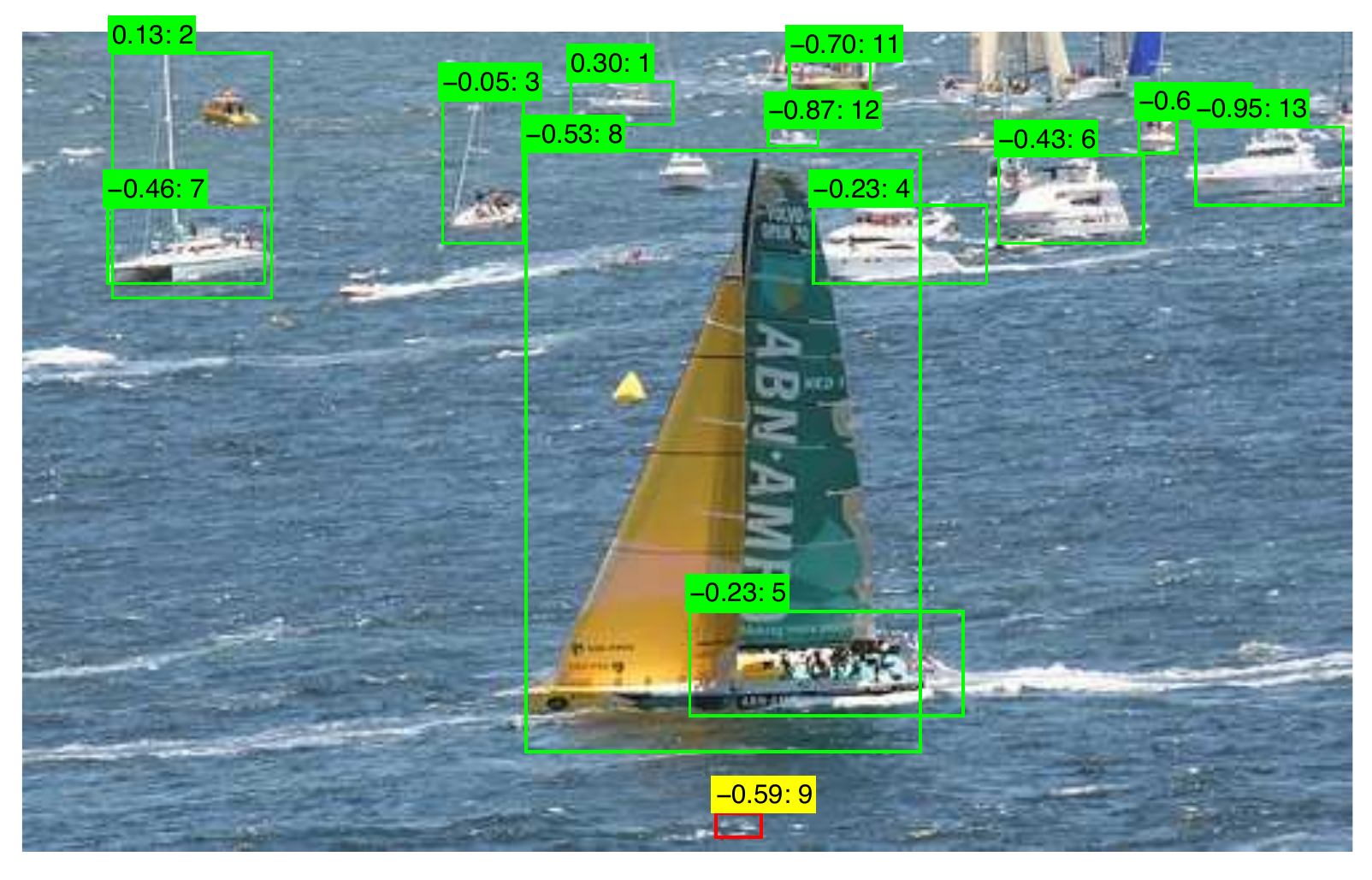}}
	
	\subfigure[R-CNN]{\includegraphics[width=0.49\textwidth]{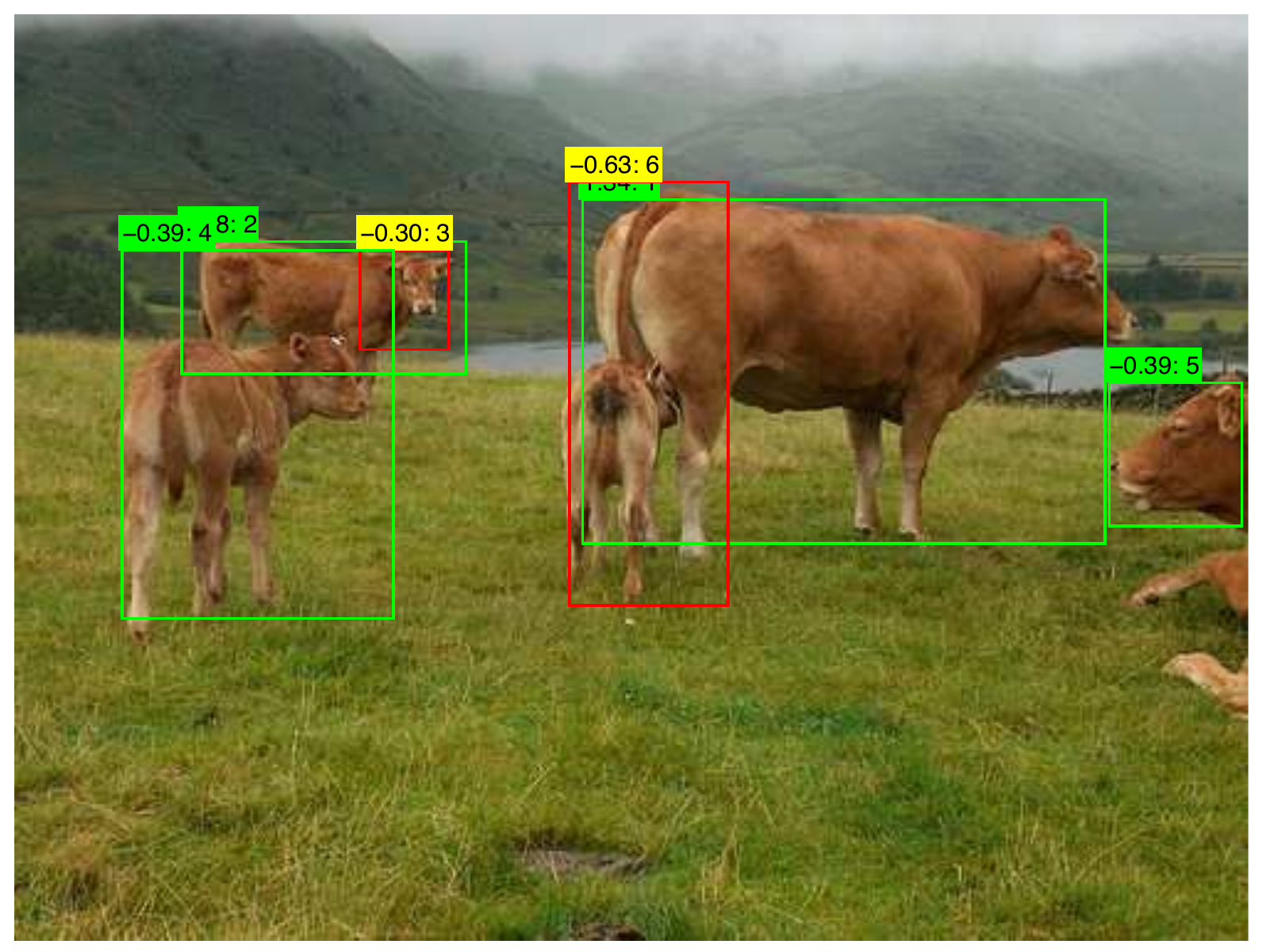}}
	\subfigure[FE\_M]{\includegraphics[width=0.49\textwidth]{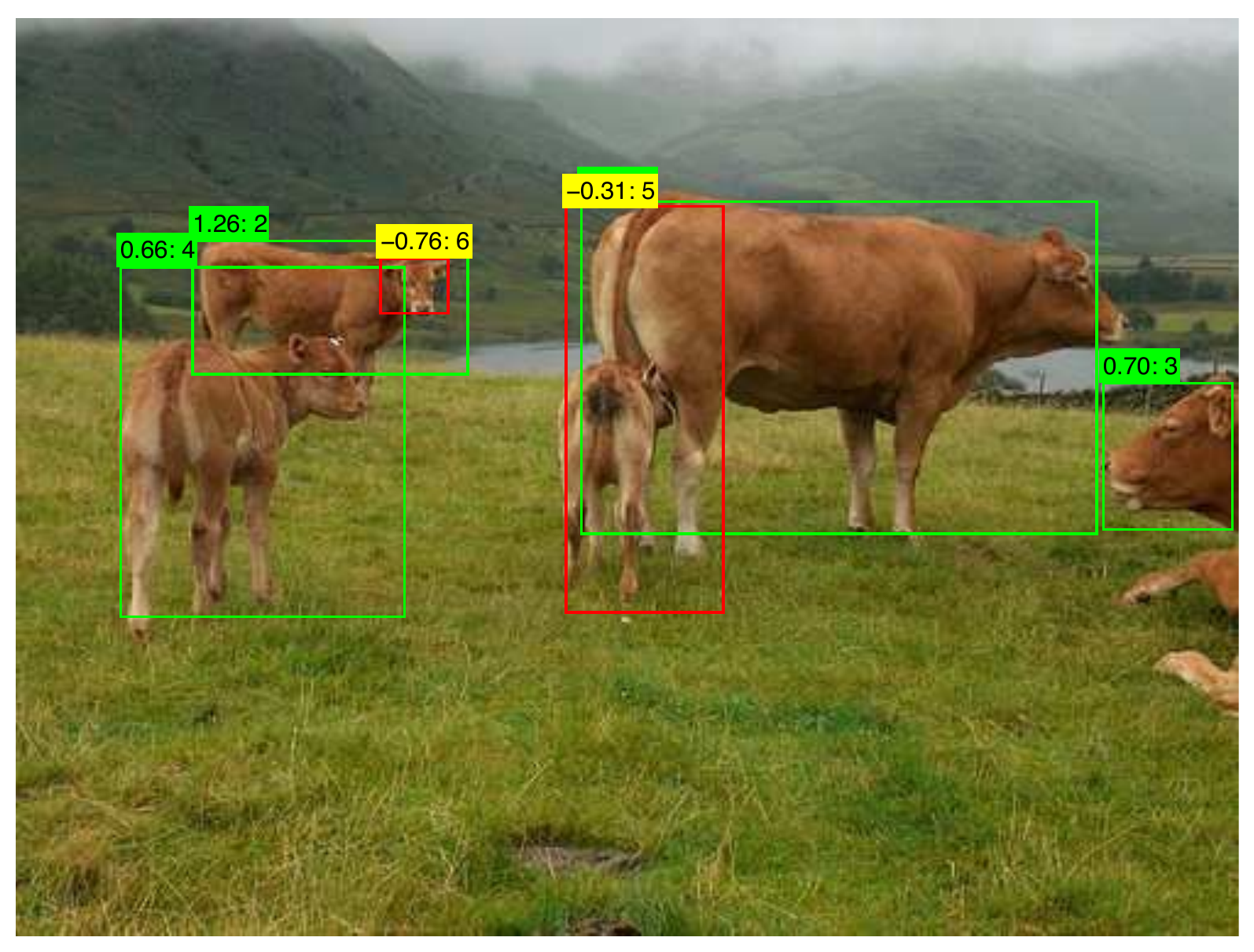}}

	\caption{Examples on VOC 2007 test set. (a) (c) in the left column are R-CNN detection results, and (b) (d) in the right column are ours. The number $score$ obtained by SVM classifier and the $rank$ of object scores in test image are shown in the top left corner of the object's bounding boxes. Green bounding boxes mean the same size and location of objects detected by both R-CNN and our method, while red bounding boxes mean detection differences appearing between R-CNN and our method.}
	\label{Images}
\end{figure*}

\subsubsection{Results Compared with R-CNN}
Compared with R-CNN, the average precision of most categories in our algorithm are obviously higher on VOC 2007, especially cow, dog and bus et al. Although a few categories are lower such as boat, bottle, and bike et al. We draw precision recall curves with four representative categories, respectively are cow, dog, boat and bottle showing in Figure~\ref{PRcurve}. For classes cow and dog, our algorithm significantly improves detection performance, but the improvements do not appear in classes boat and bottle. We find that when the object size is small, both the R-CNN and our performance are lower than those with big size. Small objects like bird and bottle are more difficult to be detected, which is also described by Russakovsky et al.~\cite{RussakovskyICCV13}.

Figure~\ref{Images} shows the R-CNN detection examples compared to ours. We show the same number of detected windows with top scores in each row. We can see that our method has few false positives (4 $vs.$ 1) in top row, and our false positives have a lower ranking (5, 6) compared to R-CNN (3, 6) in bottom row. These examples
tell us that our algorithm can train a better SVM classifier with merged training feature set.

\subsection{Drops Visualization}
 Figure ~\ref{drop} shows the drops in our algorithm. We visualize these abandoned feature channels by the introduced method in section \ref{sec2}. Two categories are shown in this paper. From this figure we can see that the correspondences of abandoned channels are aligned to those uncorrelated, tanglesome and inutile concepts which are compared to their own categories.
 
\subsection{Results on VOC 2010-2012}
The complete evaluation on VOC 2010-2012 test sets is given in Table ~\ref{2010results}~\ref{2012results}. The data sets are identical for VOC 2011 and 2012, so we just present the results of VOC 2012. The results are considerably improved on both of the two datasets. On VOC 2010, we achieve 56.4\% mAP  $vs.$ 53.7\% mAP of R-CNN and on VOC 2012, our performance is 56.3\% mAP $vs.$ R-CNN's 53.3\% mAP.

\section{Conclusions}
In this paper we consider the correspondence on a fine feature level than object category and propose an entropy-based model to drop out some negative feature elements in pool$_5$ to generate somewhat modified CNN features with very low computational complexity. A linear SVM classifier is well trained with merged CNN feature sets. 
It makes the object detection system to achieve 60.1\%, 56.4\% and 56.3\% performances on PASCAL VOC 2007, 2010 and 2012 test datasets, which are beyond all previously published results. The results indicate that our approach is much more effective with low computational cost.

{\small
\bibliographystyle{ieee}
\bibliography{myBib}
}

\end{document}